\documentclass[conference]{IEEEtran}
\IEEEoverridecommandlockouts
\usepackage{cite}
\usepackage{amsmath,amssymb,amsfonts}
\usepackage{algorithmic}
\usepackage{graphicx}
\usepackage{array}
\usepackage{tabularx}
\usepackage{subcaption}
\usepackage{textcomp}
\usepackage{xcolor}
\usepackage{booktabs}
\usepackage{microtype}
\def\BibTeX{{\rm B\kern-.05em{\sc i\kern-.025em b}\kern-.08em
    T\kern-.1667em\lower.7ex\hbox{E}\kern-.125emX}}

\usepackage{todonotes}
\usepackage{hyperref}
\usepackage{cleveref}

\newcommand{\plainAgentName}[0]{NetPlay}
\newcommand{\agentName}[0]{\texttt{NetPlay}}
\newcommand{\autoascend}[0]{\texttt{autoascend}}
\newcommand{\quotes}[1]{``#1''}

\usepackage{caption}
\begin{document}

% ToDo Find a good description for what the LLM is precisely doing in our agent
%\title{Inquirying LLMs to play NetHack\\The Potential \& Limitations of LLM Agents}
%\title{Playing NetHack with LLMs\\The Potential \& Limitations of Inquiry Agents}
%\title{Playing NetHack with LLMs\\The Potential \& Limitations of Zero-Shot Agents}
%\title{Playing NetHack with LLMs\\\huge{The Potential \& Limitations of Zero-Shot LLM Agents}}
%\title{Playing NetHack with LLMs:\\Potential \& Limitations as Zero-Shot Agents}
%\title{\textcolor{red}{(WIP)} A reactive LLM Agent for NetHack\\ Potential \& Limitations\\}
\title{Playing NetHack with LLMs:\\Potential \& Limitations as Zero-Shot Agents}

%\title{\textcolor{red}{(WIP)} LLMs in Action\\A reactive LLM Agent for NetHack\\

\author{
    \IEEEauthorblockN{Dominik Jeurissen, Diego Perez-Liebana, Jeremy Gow}
    \IEEEauthorblockA{\textit{Queen Mary University of London} \\
    \{d.jeurissen, diego.perez, jeremy.gow\}@qmul.ac.uk}
\and
    \IEEEauthorblockN{Duygu \c{C}akmak, James Kwan}
    \IEEEauthorblockA{\textit{Creative Assembly} \\
    \{duygu.cakmak, james.kwan\}@creative-assembly.com}
}

\maketitle

\begin{abstract}
Large Language Models (LLMs) have shown great success as high-level planners for zero-shot game-playing agents. However, these agents are primarily evaluated on Minecraft, where long-term planning is relatively straightforward. In contrast, agents tested in dynamic robot environments face limitations due to simplistic environments with only a few objects and interactions. To fill this gap in the literature, we present \plainAgentName{}, the first LLM-powered zero-shot agent for the challenging roguelike NetHack. NetHack is a particularly challenging environment due to its diverse set of items and monsters, complex interactions, and many ways to die.

\plainAgentName{} uses an architecture designed for dynamic robot environments, modified for NetHack. Like previous approaches, it prompts the LLM to choose from predefined skills and tracks past interactions to enhance decision-making. Given NetHack's unpredictable nature, \plainAgentName{} detects important game events to interrupt running skills, enabling it to react to unforeseen circumstances. While \plainAgentName{} demonstrates considerable flexibility and proficiency in interacting with NetHack's mechanics, it struggles with ambiguous task descriptions and a lack of explicit feedback. Our findings demonstrate that \plainAgentName{} performs best with detailed context information, indicating the necessity for dynamic methods in supplying context information for complex games such as NetHack.

\end{abstract}


\section{Introduction}
% LLMs have been successfully applied to various games (Minecraft, Crafter, Text Games, what else?)
% These approaches exploit the somewhat predictable nature of the tasks (Diamonds can always be found by diggin down and searching)
% Investigated utilizing LLMs for a more dynamic and reactive (same word?) agent
% Implemented a nethack agent that utilizes handcrafted skills to fulfill given inquiries.
% LLMs are utilized to create a very flexible agent that is able to handle the huge amount of edge-cases in NetHack.
% Resulting agent can achieve a wide range of tasks (Mini-experiments). It is also able to somewhat competently play NetHack when focusing on winning (agent comparison)
% It displays a great deal of creativity, but is still limited by (naivity?) as well as a lack of spatial awareness.
% The paper is structured as follows
% Github
Recently, agents based on Large Language Models (LLMs) \cite{llm_overview} have been successfully applied to robot environments \cite{innermonologue, llmplanner} and Minecraft \cite{jarvis, gitm, voyager}, among others. These agents do not require pre-training and typically involve prompting an LLM to solve tasks by choosing from predefined skills. They have proven effective for tasks demanding extensive knowledge, like crafting a diamond pickaxe in Minecraft. Additionally, they can understand a wide range of task descriptions.

LLM agents utilizing predefined skills are particularly promising for game development as developing a set of simple skills is often more feasible than designing an entire agent. However, existing studies predominantly focus on the capabilities of LLMs for game-playing, neglecting to address their limitations. Evaluations typically focus on predictable tasks, for example, finding a diamond in Minecraft, which can consistently be achieved through strip mining. Many games require more dynamic decision-making, where long-term planning is challenging, and the correct course of action is more ambiguous. While evaluations have been done on more dynamic robot environments, these environments often contain only a handful of objects and lack complex interactions.

We build upon existing literature by evaluating an LLM agent in the context of the complex and unpredictable roguelike NetHack \cite{NetHack}. NetHack is a challenging game with many monsters, items, interactions, partial observability, and an intricate goal condition. The sheer size of NetHack, paired with the many sub-systems the player has to understand, make it an excellent candidate for evaluating the limitations of LLM agents. NetHack's description files also allow us to define levels, enabling us to evaluate the agent's abilities in isolation.

In the following, we present \textbf{(\agentName{})}, a GPT-4 powered agent designed to tackle a wide range of tasks in NetHack. \agentName{} is inspired by \autoascend{} \cite{autoascend} a handcrafted agent that won the NetHack Challenge 2021 \cite{nle_challenge}. While \autoascend{} relied on a large network of handcrafted rules to handle the complexity of NetHack, \agentName{} only requires a set of isolated skills. Our experiments show that \agentName{} can interact with most of NetHack's game mechanics and that it excels in following detailed instructions. Additionally, the agent exhibits creative behavior when focusing its attention on a specific problem. However, when tasked to play autonomously, \agentName{} is far outperformed by \autoascend{}. Consequently, this paper delves into reasons for this, such as the agent's struggles to handle ambiguous instructions, confusing observations, and a lack of explicit feedback.

We begin in \cref{sec:background} with an overview of NetHack and a review of existing work on LLM-powered agents. \Cref{sec:approach} discusses the architecture of \agentName{}, including many of the design decisions we had to make due to limitations caused by the LLM. In \cref{sec:experiments}, we first evaluate \agentName{}'s ability to autonomously play the game and compare its performance with a simple handcrafted agent and \autoascend{}. We follow this up with an in-depth analysis of the agent's behavior across various isolated scenarios. Subsequently, we analyze the experiment results in \cref{sec:potential_and_limitations} and conclude this study in \cref{sec:conclusion}. The source code can be found on GitHub\footnote{\url{https://github.com/CommanderCero/NetPlay}}.
\section{Background} \label{sec:background}

\subsection{NetHack}
% nle, minihack, nle-language-wrapper, autoascend

\begin{figure*}[!htb]
    \centering
    \begin{subfigure}{0.65\textwidth}
        \includegraphics[width=\linewidth]{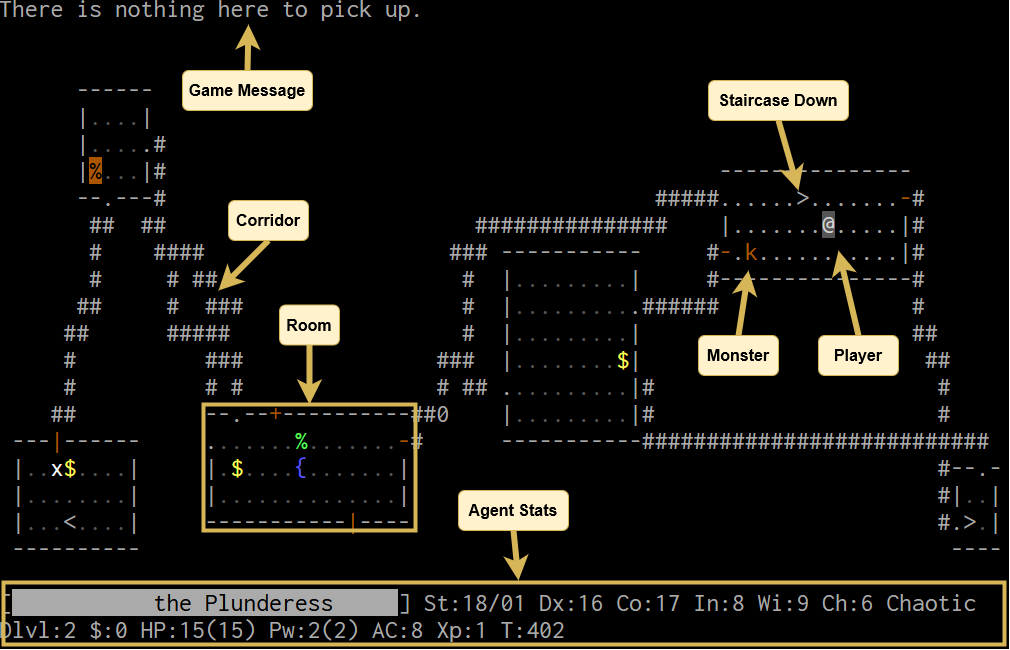}
    \end{subfigure}
    \hfill
    \begin{subfigure}{0.30\textwidth}
        \includegraphics[width=\linewidth]{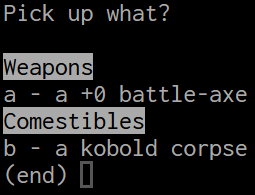}
    \end{subfigure}
    \caption{The terminal view of the game NetHack. The left image presents an annotated view of the in-game screen, featuring the game's map, an example of a game message, and the agent's stats. The right image showcases a menu for picking up items from a tile containing multiple objects. Image source: \href{https://alt.org/nethack/}{alt.org/nethack}}
    \label{fig:nethack_example}
\end{figure*}

NetHack \cite{NetHack}, released in 1987, is an extremely challenging turn-based roguelike that continues to receive updates to this date. The objective is to traverse 50 procedurally generated levels, retrieving the Amulet of Yendor and successfully returning to the surface. Doing so unlocks the final challenge of the game: the four elemental planes, followed by the astral plane, where players must present the Amulet to their deity. See \cref{fig:nethack_example} for a screenshot of the game.

Most aspects of the game are generated, such as level layouts, the player's starting class, and the inventory. The levels follow a somewhat linear structure with many branches and sub-dungeons in between. For instance, the entrance to the gnomish mines always spawns somewhere between depth 2 and 4, giving the player the option to explore them immediately or to postpone exploration until they are stronger.

NetHack encompasses a diverse array of monsters, items, and interactions. Players must skillfully utilize their resources while avoiding many of the game's lethal threats. Even for seasoned players possessing extensive knowledge of the game, victory is far from guaranteed. The game's inherent complexity requires players to continuously re-assess their situation to adapt to the unpredictability of the elements at play.

Nethack uses description files (\textbf{des-files}) to describe special levels like the oracle level that always contains a room with an oracle monster, centaur statues, and four fountains. \mbox{Des-files} offer extensive control over the level-generation process, allowing entirely handcrafted levels or a slightly constrained level-generation process.

\subsection{NetHack Learning Environment}
The NetHack Learning Environment \textbf{NLE} \cite{nle} serves as a reinforcement learning environment for playing NetHack 3.6.6. NLE offers easy access to most aspects of the game, such as the map, the agent's inventory, game messages, and the player's stats. While NLE provides simplified environments for learning purposes, it also allows users to play the entire game without any restrictions.

MiniHack \cite{minihack} utilizes NLE alongside des-files to construct small-scale environments that isolate specific challenges that agents will encounter in NetHack. Although MiniHack provides a list of challenges, its primary purpose is to streamline the process of designing new challenges.

\subsection{autoascend}
In the 2021 NeurIPS NetHack Challenge \cite{nle_challenge}, participants tackled the symbolic and neural tracks, where solutions were either handcrafted or designed using machine learning. Notably, the top-performing agents were exclusively symbolic, with the \autoascend{} agent emerging as the frontrunner \cite{nle_challenge_results}.

The \autoascend{} agent \cite{maciejsypetkowski_2024_autoascend} succeeded by meticulously parsing observations and creating an internal state representation to track essential information. The agent utilized the enriched data to implement a behavior tree by hierarchically combining strategies representing specific behaviors, like fighting, picking up objects, or exploring levels. Overall, \autoascend{}'s strategy consists of staying on the first dungeon level until reaching experience level 8, after which it will rapidly progress deeper into the dungeon. While following this general strategy, \autoascend{} uses many sub-strategies to improve its chance of success, such as a solver for solving the Sokoban levels, using altars for farming or identifying items, or dipping a long sword into a fountain to gain Excalibur.

Despite its victory, \autoascend{}'s performance  depended heavily on its starting class, demonstrating optimal results with the Valkyrie class. The agent occasionally descended to depth 10 and reached experience level 10. However, it is crucial to highlight that reaching around depth 50 is only one of the objectives to beat NetHack, emphasizing how challenging the environment still is.

\subsection{LLM Agents}
Recently, a plethora of LLM-based agents have emerged, aiming to leverage the planning capabilities of these models. A prominent testbed for these agents is Minecraft \cite{jarvis, gitm, voyager}, primarily focusing on the agent's ability to obtain the various items in the game. While the details vary, most approaches implement a closed-loop planning system in which the LLM generates a plan consisting of a sequence of predefined skills. The plan is then executed and, in case of failure, the agent will re-plan using only feedback from the previous plan. A noteworthy aspect of these agents is the storage and reuse of successful plans, significantly enhancing overall performance due to the hierarchical nature of obtaining items like a diamond pickaxe. The agents primarily utilize an LLM for their knowledge of how to acquire items. However, one agent has demonstrated the ability to construct structures with human feedback \cite{voyager}.

Other popular applications are robot environments, where tasks include rearranging objects on a tabletop, interacting within a kitchen, or engaging in simulated household activities \cite{DEPS, code_as_policies, innermonologue, llmplanner}. Because these environments require more dynamic decision-making compared to acquiring items in Minecraft, agents like DEPS \cite{DEPS} and Inner Monologue \cite{innermonologue} adopt a distinctive approach. Instead of relying solely on feedback from the last failed plan, they re-plan by considering a substantial portion of their recent interaction history. Similar to our approach, Inner Monologue models the interaction history as a chat containing the LLM's actions and thoughts, human feedback, and feedback from the environment, such as scene descriptions and if an action was successful. While the robot environments require more dynamic decision-making, the complexity of the observations is limited, usually consisting of a list of visible objects with spatial information being omitted as the low-level skills are handling it.

An alternative use of LLMs involves employing them to design reward functions, which are then used to train reinforcement learning agents \cite{eureka, motif, reward_design}. Most relevant to our work, Motif employs an LLM to learn various playstyles in NetHack. It achieves this by tasking the LLM to decide which of NetHack's game messages it prefers. Motif can leverage these preferences to learn reward functions for different playstyles by conditioning the LLM to prefer game messages associated with a specific playstyle, such as fighting monsters.
\section{NetPlay} \label{sec:approach}

\begin{figure*}[!htb]
    \centering
    \includegraphics[width=\textwidth]{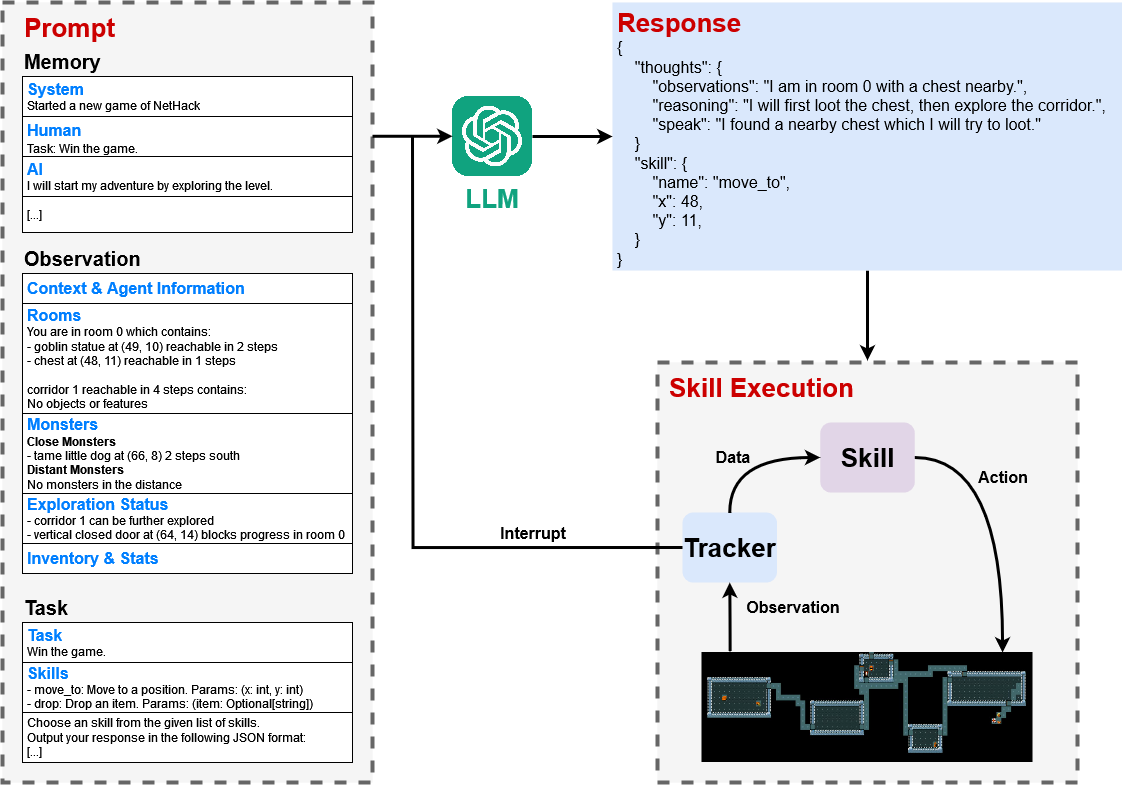}
    \caption{Illustration of \agentName{} playing NetHack. The process involves constructing a prompt using messages representing past events, the current observation, and a task description containing available skills and the desired output format. The response is parsed to retrieve the next skill. While executing the selected skill, a tracker enriches the given observations and detects important events, such as when a new monster appears. When the skill is done, or events interrupt the skill execution, the agent will restart the prompting process.}
    \label{fig:architecture}
\end{figure*}

This section discusses our LLM-powered Nethack agent \agentName{}. See \cref{fig:architecture} for an overview of the architecture. Long-term planning in NetHack proves challenging due to its unpredictability, as we cannot know when, where, or what will appear as we explore. Consequently, our agent shares many similarities with Inner Monologue, which is designed for dynamic environments. It implements a closed-loop system where the LLM selects skills sequentially while accumulating feedback in the form of game messages, errors, or manually detected events. Although we avoid constructing entire plans, the LLM's thoughts are included for future prompts, allowing for strategic planning if deemed necessary by the LLM.

\subsection{Prompting}
We prompt the LLM to choose a skill from a predefined list. The prompt comprises three components: \((a)\) the agent's short-term memory, \((b)\) a description of the observation, and \((c)\) a task description alongside the output format.

\((a)\) The observation description primarily focuses on the current level alongside additional data like context, inventory, and the agent’s stats. Because we do not use a multi-modal LLM, we attempt to convey spatial information by dividing the level into structures like rooms and corridors. Each structure is described using a unique identifier, the number of steps to reach it, the objects it contains with their respective positions, and the number of steps to reach each object. Monsters are described separately from the structures by categorizing them as close or distant, indicating their potential threat level. Each monster is described using its name, position, and number of steps to reach it. For close monsters, we also include compass coordinates. The LLM is also informed about which structures can be further explored alongside the positions of boulders and doors that block exploration progress. Note that despite our emphasis on providing spatial information, navigating the environment proved challenging for the LLM. Consequently, we automated a large portion of the exploration process using a single skill, potentially rendering certain aspects of this observation description obsolete.

\((b)\) The short-term memory is implemented using a list of messages representing the timeline of events. Each message is either categorized as system, AI, or human. System messages convey feedback from the environment like game messages or errors, AI messages capture the LLM's responses, and new tasks are indicated by human messages. Note that while it is possible for a human to provide continual feedback, we only study the case where the agent is given a task at the start of the game. The memory size is capped at 500 tokens, with older messages being deleted first. Observation descriptions are not stored in the memory due to their size.

\((c)\) The task description includes details about the current task, available skills, and a JSON output format. We employ chain-of-thought prompting \cite{chain_of_thought_prompting} to guide the LLM to a skill choice.

\subsection{Skills}
% Biggest hurdle for NetHack was to explore
% Most complex handcrafted skills have been inspired/copied by autoascend and they do exploration / movement (explore_level, go_to, move_to)
% ToDo CHECK IF I ACTUALLY DID THIS Also added fighting skill from autoascend for completion. However LLM is also able to fight with move_to alone
% Remaining skills have been kept as simple as possible to give the LLM as much freedom as possible
% Had to simplify skills somewhat as LLM kept trying to do the intuitive things (aka changed drop which will then trigger the dialogue to drop item_letter)

\begin{table*}
    \caption{\textbf{Skill Examples:} Skills represent parametrizable behaviors that the LLM uses to play the game. The name, parameters, and descriptions help to understand what each skill does. For some skills, the LLM can omit optional parameters marked in [square brackets]. Note that the skill type is only used internally and does not matter for the final agent.}
    \begin{tabular}{l | l | l | l}
        \toprule
        \textbf{Type} & \textbf{Name} & \textbf{Parameters} & \textbf{Description} \\
        \midrule
        Special & explore\_level & & Explores the level to find new rooms, as well as hidden doors and corridors. \\
        Special & set\_avoid\_monster\_flag & value: bool & If set to true skills will try to avoid monsters. \\
        Special & press\_key & key: string & Presses the given letter. For special keys only ESC, SPACE, and ENTER are supported. \\
        Position & pickup & [x: int, y: int] & Pickup things at your location or specify where you want to pickup an item. \\
        Position & up & [x: int, y: int] & Go up a staircase at your location or specify the position of the staircase you want to use. \\
        Inventory & drop & item\_letter: string & Drop an item. \\
        Inventory & wield & item\_letter: string & Wield a weapon. \\
        Direction & kick & x: int, y: int & Kick something. \\
        Basic & cast & & Opens your spellbook to cast a spell. \\
        Basic & pay & & Pay your shopping bill. \\
        \bottomrule
    \end{tabular}
    
    \label{table:skills}
\end{table*}

Skills, similar to strategies in \autoascend{}, implement specific behaviors by returning a sequence of actions. They accept both mandatory and optional parameters as input. Skills can generate messages as feedback, which are stored in the agent's memory. Messages are often used, for example, to report why a skill failed. An excerpt of skills can be found in \cref{table:skills}.

Navigation is automated through skills like \quotes{move\_to x y} or \quotes{go\_to room\_id}. However, exploring levels with only these skills proved challenging for the LLM. To address this, we introduced the \quotes{explore\_level} skill, which uses the exploration strategy from \autoascend{}. This skill explores the current level by uncovering tiles, opening doors, and searching for hidden corridors. We removed the ability to kick open doors to avoid potential issues such as aggravating shopkeepers. Note that the agent can still decide to kick open doors using a separate \quotes{kick} skill. All movement-related skills will attack monsters that are in the way. The LLM can turn off this behavior using the \quotes{set\_avoid\_monster\_flag} skill.

To indicate when the agent is done with a given task, it has access to the 
\quotes{finish\_task} skill. Additionally, the LLM is equipped with the \quotes{press\_key} and \quotes{type\_text} skills for navigating NetHack's various game menus. While a menu is open, only the \quotes{finish\_task} and text input skills remain available.

The remaining skills are thin wrappers around NetHack commands, such as drink or pickup. However, these commands often involve multiple steps, such as confirming which item to drink or first positioning the agent correctly to then pick up an item. Consequently, the LLM often assumed that the \quotes{drink} command accepts an item parameter or that \quotes{pickup} works seamlessly regardless of the agent's current position. To mitigate these issues, we implemented four types of command skills. Base commands only invoke the command. Position commands offer the option to first move to the desired location. Inventory commands accept an item parameter to resolve the following popup menu. Finally, direction commands like \quotes{kick} move the agent close to a desired position before executing the command in the correct direction.

\subsection{Agent Loop}
Upon receiving a new task, the agent prompts the LLM to select the first skill to execute. The LLM's thoughts and the selected skill are stored in the agent's memory as feedback. While executing the chosen skill, a data tracker observes and records details such as found structures, features hidden by monsters or items, which tiles the agent has already seen or searched, and events. The information collected by the data tracker is used by skills to make decisions.

The data tracker also looks for specific events in the game to provide additional feedback to the LLM. Events include new in-game messages, newly discovered structures, level changes or teleports, stat changes, low health, and the discovery of new monsters, items, and some map features such as fountains or altars.

A skill continues to run until completion or interruption. Skills are interrupted when specific events occur, such as changing the level, teleporting, discovering new objects, and reaching low health. In addition to events, many skills are interrupted when a menu shows up due to their inability to handle them. Regardless of why a skill stopped, the agent then prompts the LLM to select the next skill. The sole exception is when the \quotes{finish\_task} skill is selected, or the game has ended, at which point the agent will stop until it receives a new task.

\subsection{Handcrafted Agent}
To assess the impact of the LLM in contrast to the predefined skills, we implemented a handcrafted agent that aims to replicate the behavior of \agentName{} with the task set to \quotes{Win the Game}. The following list shows a breakdown of the agent's decision-making process.

\begin{enumerate}
\item Abort any open menu, as we did not implement a way to navigate them.
\item If there are hostile monsters nearby, fight them.
\item If health is below 60\%, try healing with potions or by praying.
\item Eat food from the inventory when hungry.
\item Pick up items, which in this case are potions and food.
\item If nothing to explore, move to the next level if possible.
\item If nothing else to do, explore the level and try kicking open doors.
\end{enumerate}

All the conditions are evaluated in sequence. Once a condition is met, a corresponding skill is executed. The selected skill will be interrupted in the same way as \agentName{}. Once a skill is interrupted, the agent will choose the next skill by again checking all conditions in order starting from the first. Note that although we aimed to imitate \agentName{}'s behavior, the provided rules are too simplistic to capture all the nuances.
\section{Experiments} \label{sec:experiments}
Our goals for the experiments were two-fold. First, to evaluate the ability of \agentName{} to play NetHack. Second, to provide an analysis of the agent's strengths and weaknesses, focusing on identifying which aspects are influenced by the LLM. 

\subsection{Setup}
All of our experiments used OpenAI's GPT-4-Turbo (\textsc{gpt-4-1106-preview}) API as LLM with the temperature set to 0 and the response format set to JSON. Other models were not considered as initial tests revealed that models like GPT-3.5 and a 70B parameter instruct version of LLAMA 2 \cite{llama2} could not correctly utilize our skills. The agent's memory size was set to 500 tokens.

The agent had access to most commands that interact with the game directly, except for some rarely relevant commands, like turning undead or using a monster's special ability. All control and system commands, like opening the help menu or hiding icons on the map, were excluded. We also implemented a time limit of 10 LLM calls, at which point the experiment would terminate if the in-game time did not advance.

\subsection{Full Runs}
\begin{table*}[!htb]
    \centering
    \caption{\textbf{Results summary} of the mean and standard error for the agents achieved score, depth, experience level, and game time.}
    \begin{tabular}{l | l | l | l | l}
        \toprule
        \textbf{Metric} & \textbf{\agentName{} (Unguided)} & \textbf{\agentName{} (Guided)} & \textbf{autoascend} & \textbf{handcrafted} \\
        \midrule
        Score & 284.85 $\pm$ 222.10 & 405.00 $\pm$ 216.38 & \textbf{11341.94 $\pm$ 11625.39} & 250.24 $\pm$ 159.17 \\
        Depth & 2.60 $\pm$ 1.39 & 2.00 $\pm$ 1.05 & \textbf{4.01 $\pm$ 3.04} & 2.35 $\pm$ 0.93 \\
        Level & 2.40 $\pm$ 1.23 & 3.30 $\pm$ 0.95 & \textbf{3.34 $\pm$ 7.69} & 2.39 $\pm$ 1.05 \\
        Time & 1292.10 $\pm$ 942.74 & 2627.40 $\pm$ 1545.12 & \textbf{21169.81 $\pm$ 9155.59} & 1306 $\pm$ 924.17 \\
        \bottomrule
    \end{tabular}
    \label{tab:performance_results}
\end{table*}

We started evaluating \agentName{} by letting it play NetHack without any constraints, tasking it to win the game. We will refer to this agent as the \quotes{unguided agent.} Although the task was to play the entire game, the agent occasionally confused its own objectives with the assigned task, resulting in the agent marking the task as done too early. To address this issue, we disabled the \quotes{finish\_task} skill for this experiment.

Due to budget limitations, we evaluated all agents using only the Valkyrie role, as most agents performed best with this class during the NetHack 2021 challenge. We conducted 20 runs with the unguided agent. Additionally, we performed 100 runs each with \autoascend{} and the handcrafted agent for comparison. After evaluating the unguided agent, we carried out an additional 10 runs employing a \quotes{guided agent} who was informed on how to play better. A detailed description of the guided agent will be provided below. For now, a summary of the results can be found in \cref{tab:performance_results}.

\Cref{tab:performance_results} shows that \autoascend{} far outperforms both \agentName{} and the handcrafted agent. While \agentName{} managed to beat the handcrafted agent by a small margin, it is likely that with a few tweaks, the handcrafted agent can also outperform \agentName{}.

The unguided agent primarily failed due to timeouts, followed by deaths caused by eating rotten corpses, fighting with low health, or being overwhelmed by enemies. Many timeouts were caused by the agent attempting to move past friendly monsters, such as a shopkeeper. By default, bumping into monsters attacks them, but for passive monsters, the game prompts the player before initiating an attack. The agent's refusal to attack these monsters often leads to a loop of canceling the prompt and moving, resulting in eventual timeouts. A similar loop took place when the agent attempted to pick up an item with a generic name on the map but a detailed name in the game's menu. This confusion led the agent to repeatedly close and reopen the menu, unable to locate the desired item.

Based on the results of the unguided agent, we constructed a guide that included strategies from \autoascend{}, such as staying on the first two dungeon levels until reaching experience level 8, consuming only freshly slain corpses to avoid eating rotten ones, and leveraging altars to acquire items. Furthermore, we provided tips for common mistakes by the unguided agent, such as avoiding getting stuck behind passive monsters and informing the agent about the time limit to avoid timeouts.

The guided agent often managed to stay alive longer by consuming freshly killed corpses and praying when hungry or at low health. Its causes of death have been a mixture of timeouts, starvation, and dying in combat. Most of the timeouts stemmed from a bug with our tracker, which fails to detect when an object disappears while being obscured by a monster. For example, the agent repeatedly attempted to pick up a dagger already taken by its pet due to the tracker's misleading observation. Despite receiving game messages indicating the absence of the item, the agent failed to recognize the situation accurately.

Because we tasked the guided agent to stay on the first two dungeon levels, its average depth is lower than that of the unguided agent. However, because monsters keep spawning over time, staying on the first levels is an excellent way to grind experience. This results in the guided agent gaining more experience than the unguided agent. Nevertheless, the agent's tendency to stay on the first dungeon levels frequently caused it to die of starvation due to not finding enough monster corpses to eat. Note that \autoascend{} had a similar starvation issue. 

\subsection{Scenarios}
After conducting the full runs, we hypothesized that although \agentName{} can be creative and interact with most mechanics in the game, it tends to fixate on the most straightforward approach for a given task. To confirm this hypothesis, we constructed various small-scale scenarios using des-files and a corresponding task description. Note that we excluded the handcrafted agent and \autoascend{} for this experiment as they cannot easily alter their behavior.

The tested scenarios evaluated \agentName{}'s ability to interact with game mechanics, follow instructions, and its creativity. We conducted five runs for each scenario, with all roles and the \quotes{finish\_task} skill enabled. We also repeated some scenarios where the agent performed poorly with additional guidelines. We censored the word NetHack for the scenarios to evaluate the agent's ability independently of its knowledge about the game. To avoid the agent never using the \quotes{finish\_task} skill, we set a time limit of 500 timesteps for creative scenarios and 200 for the others. See \cref{table:scenarios} for a summary of the tested scenarios and their results.

The tested scenarios show that \agentName{} performs best when provided with concrete instructions. The \textsc{focused boulder} task and both \textsc{escape} tasks, in particular, highlight how the agent can act creative if we focus its attention on a specific problem. However, without very detailed instructions, the agent often fails to do what it wants due to incorrect actions and a lack of explicit feedback.

The agent's struggle with explicit feedback is particularly evident in the \textsc{bag} and \textsc{multipickup} scenarios, where the agent often failed to navigate the menus correctly. While it understood the menus and often chose the correct course of action, it often failed by forgetting a crucial step, such as closing the menu.

\begin{table*}[htbp]
    \centering
    \renewcommand{\arraystretch}{1.4}
    \caption{\textbf{Scenarios:} A detailed description of all the tested scenarios, their results, and the agent's success rate. Note that in some scenarios, the agent did not use the \quotes{finish\_task} skill, even after completing it. We still count these as success.}
    \resizebox{\textwidth}{!}{
    \begin{tabularx}{\textwidth}{|c|c|X|X|}
        \hline
        \textbf{Scenario} & \textbf{Success} & \textbf{Description} & \textbf{Results} \\
        \hline
        \multicolumn{4}{|c|}{\textbf{Game Mechanics}} \\
        \hline
        
        \textsc{bag} & 1/5 & A room with four random objects and a bag of holding with the task of stuffing all objects into the bag. & The bag of holding menu is quite complex. The agent was only successful when using the option that automatically stuffs all items into the bag. In the other cases, the agent forgot to mark an item or to confirm its selection. \\
        
        \textsc{guided bag} & 3/5 & Same as \textsc{bag}, but we told the agent the quickest way to pick up items and to navigate the bag's menu. & The agent used the automatic option three times. In the other cases, the agent marked the task done too early, stating that it would pick up the remaining items next. \\
        
        \textsc{multipickup} & 3/5 & A room with 2-5 objects on the same spot, challenging the agent to navigate the multipickup menu. & The agent often picked up items inefficiently by opening the pickup menu multiple times. It failed twice by forgetting to confirm its item selection. \\
        
        \textsc{wand} & 1/5 & A room with a statue and a wand with the task of hitting the statue with the wand. & The agent often failed by standing atop the statue and casting the wand onto itself. Only once did the wand spawn next to the statue, causing the agent to cast the wand towards the statue. \\
        
        \textsc{guided wand} & 5/5 & Same task as \textsc{wand}, but we asked the agent to stand next to the statue instead of on top of it and fire in the statue's direction. & Most of the time, the agent succeeded on the first try, except once when he got it on the second try after repositioning himself. \\
        
        \hline
        \multicolumn{4}{|c|}{\textbf{Instructions}} \\
        \hline
       
        \textsc{ordered} & 5/5 & A room with the task to pick up two wands, then a scroll of identification, and finally to identify one wand. & The agent executed the tasks accurately in the given order. \\
        
        \textsc{unordered} & 3/5 & A room with the task to drink from a fountain, open a locked and a closed door, and kill a monster in any order. & The agent completed the tasks in no particular order. One fail stemmed from high-level mobs spawning from the fountain, and one from incorrectly using the lockpick. \\
        
        \textsc{alternative} & 5/5 & Three rooms with a fountain and a potion somewhere. The task was to drink from a fountain or a potion. & The agent always drank from the fountain, which in all cases was found first or was closest to the agent. \\
        
        \textsc{conditional} & 4/5 & Three rooms, with only a single potion hidden in one of the rooms. The task was to drink from a fountain, or if unavailable a potion. & The agent always drinks the first potion it finds without exploring further. In one case, it deemed the task impossible due to spawning with no fountain or potion in sight. \\
        
        \hline
        \multicolumn{4}{|c|}{\textbf{Creativity}} \\
        \hline
        
        \textsc{carry} & 1/5 & The agent has to carry two very heavy objects through a monster-filled room. We also provided tools such as a bag of holding, a teleportation wand, and an invisibility cloak. & The agent often refused to play because it could not see the required items or it dropped them in the wrong room. \\
        
        \textsc{guided carry} & 4/5 & Same task as \textsc{carry}, but we told the agent to prioritize killing monsters first, to carry only one of the heavy items at a time, and to use the teleportation wand for easier travel. & Most of the time, the agent carried only one item, and it often used the wand to teleport. It failed once by dropping one item in the incorrect room. \\
        
        \textsc{boulder} & 1/5 & Two rooms connected by a corridor with a boulder. The agent starts either with pickaxes or wands to remove the boulder. & When given only wands, the agent only used explore\_level and ignored the boulder. Only once did it start with a pickaxe that it used to mine the boulder. \\
        
        \textsc{focused boulder} & 3/5 & Same task as \textsc{boulder}, but the agent was told to remove any boulders blocking its path. & The agent often tried kicking the boulder, which failed, after which it then
used a pickaxe or a wand. It failed twice due to not correctly utilizing the available tools. \\
        
        \textsc{guided boulder} & 5/5 & Same task as \textsc{focused boulder}, but the agent was told explicitly to remove the boulder with the wands or pickaxes. We also provided directions on how to utilize the tools. & In all cases, the agent quickly used the pickaxe or a wand to remove the boulder. \\
        
        \textsc{escape} & 3/5 & The agent must escape from a stone-walled room. Escape methods: Digging with a wand through a wall, teleporting with a wand, or morphing into a wall-phasing monster using a polymorph control ring with a polymorph wand. & 
The agent escaped twice by teleporting, despite initial teleport failure. It also experimented with the wand of digging, casting it in all directions to find an exit. It failed twice due to incorrectly using the wands. \\

        \textsc{hint escape} & 5/5 & Same as \textsc{escape}, with a hint engraved on the floor. The hint either reveals which wall is brittle and leads to an escape or hints at the name of the wall-phasing monster. & After finding the hint, the agent often used the suggested escape method, except for one occasion when it teleported instead. In one instance, the initial attempts to dig through the wall failed, so it resorted to exploring other methods.\\
        
        \hline
    \end{tabularx}
    }
    \label{table:scenarios}
\end{table*}

\section{Potential and Limitations} \label{sec:potential_and_limitations}
\agentName{} uses a similar architecture to Inner Monologue and DEPS, which have shown promising results for simple dynamic environments. Our experiments show that despite the immense complexity of NetHack, the agent can fulfill a wide range of tasks given enough context information. To our knowledge, this is the first NetHack agent to exhibit such flexible behavior. However, the benefits of the presented approach seem to diminish the more ambiguous a given task is, making tasks such as \quotes{Win the Game} impossible.

A promising use case of the presented architecture is regression testing during game development. Game developers could test specific aspects of their game by providing \agentName{} with detailed instructions on what to test. This approach could not only streamline the testing process, but it would also benefit from \agentName{}'s flexibility, enabling the tests to adapt dynamically as the game evolves.

Given \agentName{}'s proficiency when given detailed context information, an obvious extension to our approach would be granting the agent access to the NetHack Wikipedia. This could be done using a skill that accepts a query and adds the resulting information to the agent's short-term memory. While we think this can improve the results at the cost of more LLM calls, finding the most relevant information for a given situation would be tricky. Instead, we recommend investing future research into automated methods for finding relevant context information, with a particular focus on finding the most successful past interactions as guidelines on how to play. 

A significant limitation of our approach lies in the predefined skills and the observation descriptions, which struggle to encompass the vast complexity of NetHack. Designing the agent to handle all potential edge cases proved challenging, as it is difficult to anticipate every scenario. 
While the premise of this approach is that the LLM can handle these edge cases, this is only true as long as we have a comprehensive description of the environment and flexible skills. In practice, achieving such a well-designed agent requires an ever-growing repertoire of skills and an observation description that grows infinitely. As such, another promising research direction is to use machine learning to replace the handcrafted components of the agent.
\section{Conclusion} \label{sec:conclusion}
In this work, we introduce \agentName{}, the first LLM-powered zero-shot agent for the challenging roguelike NetHack. Building upon an existing approach tailored for simpler dynamic environments, we extended its capabilities to address the complexities of NetHack. We evaluated the agent's performance on the whole game and analyzed its behavior using various isolated scenarios.

\agentName{} demonstrates proficiency in executing detailed instructions but struggles with more ambiguous tasks, such as winning the game. Notably, a simple rule-based agent can achieve comparable performance in playing the game. \agentName{}'s strength lies in its flexibility and creativity. Our experiments show that, given enough context information, \agentName{} can perform a wide range of tasks. Moreover, by focusing its attention on a particular problem, \agentName{} is adept at exploring a wide range of potential solutions but often with limited success due to a lack of explicit feedback guiding it.

\section{Acknowledgements}
% IGGI
This work was supported by the EPSRC Centre for Doctoral Training in Intelligent Games \&
Games Intelligence (IGGI) (EP/S022325/1).

% Creative Assembly
This work was supported by Creative Assembly.

\bibliography{references}
\bibliographystyle{IEEEtran}

% No space :(
%\section*{Appendix}
%\subsection*{All mini experiments}
%\subsection*{Example prompts}
%\subsection*{BiModal Test}

\end{document}